\documentclass[10pt]{article}

\usepackage[utf8]{inputenc}
\usepackage[T1]{fontenc}
\usepackage{microtype}
\usepackage{times}
\usepackage{graphicx}
\usepackage{booktabs}
\usepackage{multirow}
\usepackage{amsmath}
\usepackage{amssymb}
\usepackage{xcolor}
\usepackage{url}
\usepackage[colorlinks=true,
            linkcolor=blue,
            citecolor=blue,
            urlcolor=blue]{hyperref}
\usepackage{cleveref}
\usepackage{enumitem}
\usepackage{caption}
\usepackage{subcaption}
\usepackage{array}
\usepackage{titlesec}
\usepackage{balance}
\usepackage{mdframed}
\usepackage{longtable}

\usepackage[
  a4paper,
  top=2.4cm,
  bottom=2.4cm,
  left=3.0cm,
  right=3.0cm
]{geometry}

\titleformat{\section}
  {\normalfont\large\bfseries}{\thesection.}{0.6em}{}
\titleformat{\subsection}
  {\normalfont\normalsize\bfseries}{\thesubsection}{0.6em}{}
\titleformat{\subsubsection}
  {\normalfont\normalsize\itshape}{\thesubsubsection}{0.6em}{}
\titleformat{\paragraph}[runin]
  {\normalfont\normalsize\bfseries}{}{0em}{}[.\enspace]
\titlespacing{\section}{0pt}{10pt plus 2pt minus 2pt}{4pt}
\titlespacing{\subsection}{0pt}{8pt plus 2pt minus 2pt}{3pt}
\titlespacing{\paragraph}{0pt}{4pt}{0pt}

\mdfdefinestyle{promptbox}{%
  linecolor=black,
  linewidth=1pt,
  backgroundcolor=white,
  innerleftmargin=10pt,
  innerrightmargin=10pt,
  innertopmargin=8pt,
  innerbottommargin=10pt,
  frametitlebackgroundcolor=black!85,
  frametitlefont=\bfseries\small\color{white},
  frametitlerule=false,
  frametitleaboveskip=4pt,
  frametitlebelowskip=6pt,
}

\newcommand{\modelname}{\textsc{ExtractConf}}
\newcommand{\hunter}{\textit{Hunter}}
\newcommand{\mapper}{\textit{Mapper}}

\begin{document}

\begin{center}

\vspace*{0.5cm}

{\LARGE\bfseries
  Beyond Logprobs: A Multi-Signal Confidence Engine\\[6pt]
  for LLM-Based Document Field Extraction%
  \footnote{Extended version of a paper accepted (Oral) at the
  RobustifAI Workshop, IJCAI-ECAI 2026, Bremen, Germany.}
}

\vspace{0.5em}
{\color{black}\rule{\textwidth}{1.8pt}}
\vspace{0.3em}

{\normalsize
  \textbf{Nitesh Kumar}$^{1}$

  \vspace{0.4em}
  $^{1}$Perfios Software Solutions Pvt. Ltd.\\
  \texttt{nitesh.kumar@perfios.com}
}
\vspace{0.5em}
{\color{black}\rule{\textwidth}{0.5pt}}
\vspace{1.0em}

\begin{center}
\begin{minipage}{0.85\textwidth}
\begin{center}
{\normalsize\textbf{Abstract}}
\end{center}
\vspace{4pt}
{\small\setlength{\parindent}{0pt}\setlength{\parskip}{5pt}
In high-stakes document processing pipelines,
including financial reconciliation, compliance verification, and procurement
automation, an LLM extraction that is silently wrong is more
dangerous than one that is visibly absent.
The central challenge is not extraction accuracy alone but
\emph{reliable confidence estimation}: knowing, field by field,
whether a given extraction can be trusted for automation or
must be deferred to human review.
Existing approaches fall short: token-level log-probabilities,
verbalized confidence, and multi-sample self-consistency all
collapse toward all-positive behaviour at practical thresholds,
offering no reliable separation between trustworthy and
untrustworthy extractions.

We present \modelname{}, a cross-domain, field-agnostic
confidence engine that grounds reliable confidence estimation in two
structurally different readings of the same document.
A field-guided \hunter{} call extracts each field
independently under schema-slot completion prompt;
a document-guided \mapper{} call scans the document holistically
and surfaces candidate values grounded in what the document
actually contains, without being anchored to any single field.
This output-structure asymmetry gives them different failure
modes: \hunter{} tends to hallucinate values for ambiguous
or absent fields, while \mapper{} tends to miss fields that are
visually non-salient. Their disagreement is independently
informative.
\modelname{} fuses signals derived from this cross-call
disagreement, LLM-internal uncertainty, OCR, image quality,
and spatial layout into a binary classifier
that operates without any domain-specific rules or retraining
for new document types or field sets.
On DocILE (55-field invoices, 26\% natural failure rate),
\modelname{} achieves 0.928~ROC~AUC and reduces selective
prediction risk by 70\% over logprob-mean.
In the top confidence band, accuracy reaches 99.1\%, a
25.8-point improvement over the base rate, enabling a practical
human-in-the-loop workflow: automate high-confidence
predictions, route the rest.
Zero-shot transfer to CORD receipts achieves 0.858~AUC without
any retraining; lightweight Lasso recalibration reduces ECE
by 89\% and Brier by 43\%, confirming the reliability signals
generalise across document domains.
}
\end{minipage}
\end{center}

\vspace{1.2em}
{\color{black}\rule{\textwidth}{0.5pt}}
\vspace{1.0em}

\end{center}

\section{Introduction}
\label{sec:intro}

Automated extraction of structured fields from business
documents is a high-value production task where incorrect
extractions carry direct financial consequences.
Recent large multimodal LLMs have substantially advanced
document information extraction~\cite{perot2024lmdx,simsa2023docile},
achieving strong accuracy on structured benchmarks spanning
invoices, receipts, and forms.
Yet production deployments require more than accuracy: they
require \emph{knowing when to trust an extraction}.
A calibrated confidence score enables workflows to route
low-confidence predictions to human review, reducing costly
downstream errors while maximising automation throughput.
This selective prediction problem has been well studied in
classification settings~\cite{geifman2017selective,corbiere2019addressing},
but remains largely unsolved for LLM-based document extraction.

The standard approach is to use token-level log-probabilities
as a proxy for correctness.
We show this fails substantially: on DocILE, a 55-field invoice
benchmark where frontier LLMs fail on 26\% of fields,
logprob-mean achieves only 0.705~ROC~AUC and degrades to an
all-positive classifier at any practical threshold.
Verbalized self-assessed confidence fares similarly (0.692~AUC).
Even self-consistency across five sampled calls
achieves only 0.744~AUC, marginally better, but with
AURC~0.138, barely below logprob-mean's~0.145, and at
5$\times$ the single-call API cost.

The failure is structural.
These methods measure confidence in the \emph{generated token
sequence}, but extraction errors in document processing are
frequently caused by failures the model cannot observe:
unreadable source material, ambiguous layouts, OCR noise.
A frontier LLM confidently transcribing OCR noise produces
high log-probabilities for a wrong answer.

\modelname{} takes a different approach: construct a reliability
signal from multiple heterogeneous sources, namely what the model
generated, what the document's image quality reveals, where
the extraction was spatially located, what the document OCR says and whether two
structurally distinct extraction calls agree.
The design question that motivates the architecture is:
\emph{what if we read the same document twice, in two ways
that have different failure modes?}

\paragraph{Contributions}
\begin{enumerate}[leftmargin=*,topsep=2pt,itemsep=2pt]
  \item We introduce a dual-call \hunter{}--\mapper{} design
    where two structurally asymmetric LLM calls approach the
    same document from opposite output-structure orientations.(\S\ref{sec:architecture}).
  \item We fuse this cross-call signal with LLM-internal
    uncertainty, document quality, and spatial layout into a
    gradient-boosted confidence classifier, achieving
    0.928~ROC~AUC and reducing selective prediction risk by
    70\% over logprob-mean (\S\ref{sec:main}).
  \item We show that all three baselines fail as routing
    signals; self-consistency costs 5$\times$ per-field yet
    yields negligible AURC improvement; and document quality
    alone  outperforms LLM-internal uncertainty
    alone. (\S\ref{sec:main}).
  \item Post-hoc recalibration (M7/M8) reduces ECE by up to
    83\%, establishing three deployment configurations for
    discrimination, calibration, and routing quality
    (\S\ref{sec:main}).
  \item Zero-shot Lasso recalibration on CORD achieves ECE
    $-$89\% and Brier $-$43\%, confirming the reliability
    signals generalise across document domains
    (\S\ref{sec:generalization}).
\end{enumerate}

\section{Related Work}
\label{sec:related}

\paragraph{Multi-signal KIE confidence}
HYCEDIS~\cite{nguyen2022hycedis} is the closest predecessor
to \modelname{}: it fuses CRNN OCR logits, LSTM lingual
features, Graph-KV structural logits, and a VAE image anomaly
score through a learned binary classifier to predict whether
each KIE extraction is correct, reporting AUC and ECE on
SROIE and CORD.
\modelname{} inherits the multi-signal fusion paradigm but
extends it to the LLM era in three directions: LLM logprobs
and entropy replace CRNN logits; the asymmetric
\hunter{}--\mapper{} design adds a cross-call signal
structurally absent from HYCEDIS; and evaluation uses DocILE,
a 55-field benchmark with 26\% natural failure rate vs.\
SROIE's 4\%, providing richer negative supervision.

\paragraph{Selective prediction and abstention}
\modelname{} is fundamentally a selective prediction system:
it decides whether to act on an extraction or defer to human
review, and is evaluated with AURC, a selective prediction
metric.
Selective prediction with neural networks was formalised by
Geifman and El-Yaniv~\cite{geifman2017selective}, who showed
that a coverage-accuracy trade-off is achievable by abstaining
on low-confidence predictions.
Corbière et al.~\cite{corbiere2019addressing} proposed
learning a separate confidence score for abstention rather
than relying on softmax probability, motivating our
CatBoost meta-classifier design.
The risk-coverage curve and AURC used throughout this paper
follow the selective prediction evaluation framework of
Geifman and El-Yaniv~\cite{geifman2019selectivenet}.
Conformal prediction~\cite{rombach2026conformal} provides
coverage guarantees post-hoc but produces set-valued outputs
rather than calibrated scalar scores; \modelname{} is
complementary, providing the graded probability estimates
that enable continuous routing decisions.

\paragraph{Ensemble disagreement and epistemic uncertainty}
The \hunter{}--\mapper{} design belongs to a broader family
of disagreement-based uncertainty estimators.
Query-by-committee~\cite{seung1992query} uses disagreement
among ensemble members as an active learning signal;
co-training~\cite{blum1998combining} exploits disagreement
between two views of the same instance to propagate labels.
Lakshminarayanan et al.~\cite{lakshminarayanan2017simple}
show that deep ensembles produce better-calibrated uncertainty
than single models via prediction disagreement.
\modelname{}'s \hunter{}--\mapper{} pair differs from these
in a critical respect: the two calls are not trained
differently nor independently initialised; they are the
same model receiving \emph{structurally asymmetric prompts}
with different failure modes.
The disagreement is therefore epistemic in origin, measuring
whether the document provides sufficient grounding
for both readings to converge, rather than reflecting parametric
variance across ensemble members.

\paragraph{LLM uncertainty quantification}
Self-consistency~\cite{wang2022self} samples multiple
generations and uses majority agreement as confidence; we
adopt this as baseline B3.
Its routing utility is limited by score granularity: with
5 calls.
BSDetector~\cite{chen2024bsdetector} and
CONSTRUCT~\cite{goh2026construct} extend consistency-based
UQ to structured extraction outputs but are
document-unaware; they use no OCR, spatial, or image
quality signals.
Semantic entropy~\cite{farquhar2024semantic} clusters
semantically equivalent generations and measures cluster
entropy; P(True)~\cite{kadavath2022language} uses LLM
self-evaluation; and verbalized
confidence~\cite{tian2023ask} elicits numeric scores.
These methods are included as model-only baselines (B1, B2).
Semantic entropy is not directly applicable to structured
field extraction, where correctness is determined by
exact or near-exact string match against ground truth,
not semantic equivalence between free-form generations.

\paragraph{Multimodal grounding uncertainty}
\modelname{}'s core insight, that extraction uncertainty
should arise from document-side grounding instability rather than
just token probability, connects to the emerging literature
on multimodal hallucination.
Li et al.~\cite{li2023evaluating} show that vision-language
models hallucinate objects not present in images with high
confidence, exactly the failure mode \hunter{} exhibits on
absent fields.
The document AI-specific variant of this problem, where OCR
confidence serves as a proxy for grounding quality, is validated
by ConfBERT~\cite{hemmer2024confbert}, which injects OCR
confidence into BERT for post-OCR error detection.

\paragraph{Calibration for structured prediction}
Jagannatha and Yu~\cite{jagannatha2020} benchmark temperature
scaling, MC-Dropout, and isotonic regression for NER and
relation extraction calibration, establishing isotonic
regression as a strong post-hoc calibrator for structured
NLP tasks, motivating our M7 design.
Nixon et al.~\cite{nixon2019measuring} provide the
multi-bin ECE framework; Kull et al.~\cite{kull2019beyond}
provide the Brier score decomposition into calibration and
refinement components used to interpret our 
mapper contribution.

\section{The \modelname{} Architecture}
\label{sec:architecture}

\begin{figure}[t]
\centering
\includegraphics[width=\textwidth]{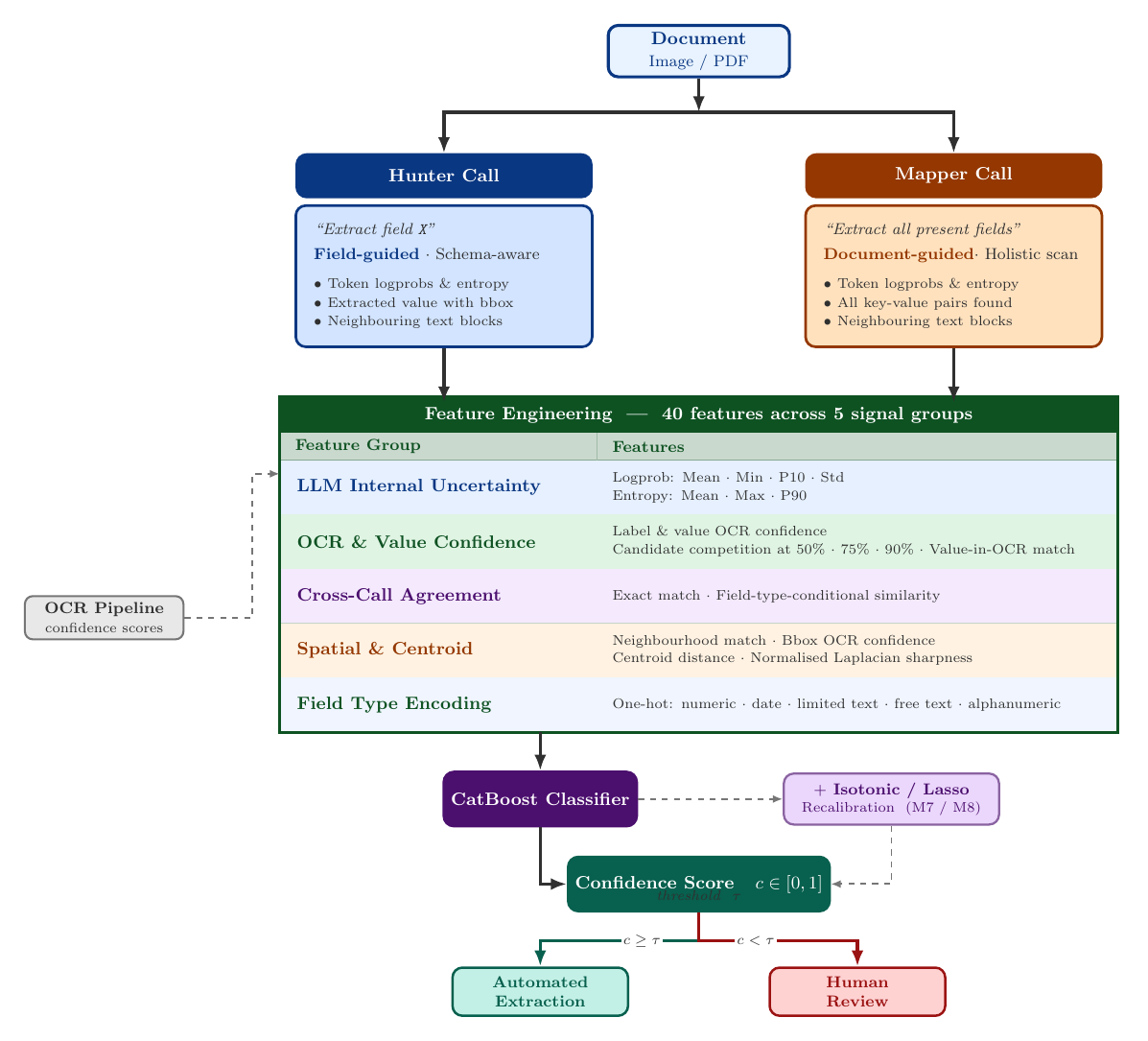}
\caption{%
\modelname{} architecture.
A document feeds two asymmetric LLM calls in parallel:
a field-guided \hunter{} (blue) and a document-guided
\mapper{} (orange).
Both report extracted values, token logprobs, and the
neighbouring text they attended to.
Value agreement, neighbourhood text overlap, and spatial
centroid divergence (from OCR bounding boxes) form the
cross-call reliability signals.
OCR confidence and image quality feed the feature table via
dashed arrows.
All 40 features are fused by a CatBoost classifier;
optional post-hoc recalibration (M7/M8) produces calibrated
probabilities.
The score $c\in[0,1]$ routes to automation or human review
at threshold~$\tau$.}
\label{fig:architecture}
\end{figure}

\subsection{Dual-Call Hunter--Mapper Design}

\textbf{\hunter{}} receives a \emph{field-guided} prompt:
given the document, extract the value of field $f$.
It returns the extracted value, per-token log-probabilities,
and the neighbouring text it attended to in the document.

\textbf{\mapper{}} receives a \emph{document-guided} prompt:
given the document, holistically scan and identify the most
salient candidate values for each target field, anchored to
what the document actually contains rather than to any single
field in isolation.
It returns a flat list of field-value pairs grounded in
document content, logprobs, and the neighbouring text each
extraction attended to.

The design rationale is the \emph{asymmetry of output
structure}.
\hunter{} fills a fixed schema slot for every field under
completion pressure; it will produce an answer regardless
of whether the field is clearly present in the document,
making it prone to confabulating plausible values for absent
or ambiguous fields.
\mapper{} produces a free evidence list of whatever it finds
most salient in the document; it reports only what is
visually grounded, making it reliable when it does surface
a value.
Their disagreement is therefore independently informative:
value clashes signal genuine extraction ambiguity, while
\hunter{} reporting when \mapper{} is silent signals a
field that required completion pressure to surface, warranting
lower confidence regardless of logprob quality.
Neither failure mode can be observed by resampling either
call, and the two-call cost is fixed per document regardless
of schema size.

\subsection{Feature Groups}
\label{sec:features}

\paragraph{Group 1. LLM Internal Uncertainty}
For both \hunter{} and \mapper{} calls independently, we
extract token-level statistics from the logprob sequence
returned by the API.
For log-probabilities: mean, minimum, 10th-percentile (P10),
and standard deviation.
For Shannon token entropy: mean, maximum, and 90th-percentile
(P90).
These statistics summarise how confident the model was over the
entire decoding trajectory, while Shannon entropy specifically
captures how dispersed the probability mass was at each step,
providing a complementary view of uncertainty beyond raw
logprobs.
This yields 14 features (7 per call), keeping the two sets
separate since \hunter{} and \mapper{} have structurally
different output formats and their uncertainty distributions
are not directly comparable.

\paragraph{Group 2. OCR Grounding and Value Confidence}
The LLM returns the extracted value and field label as it read them from
the document. We ground these against the document OCR
token list to compute region-level confidence scores.

\textit{Value-in-OCR match} is a binary indicator of
whether the extracted value appears exactly anywhere in the raw OCR
text of the document.
This is computed for both calls, detecting hallucinated
values that have no grounding in any OCR token regardless
of field assignment.

\textit{Candidate competition features} capture grounding
evidence without requiring a single exact match. It tries to capture how much alternative OCR evidence also looks compatible.
For the \hunter{} label region, we compute the fraction of
OCR candidates whose similarity to the extracted label
exceeds three thresholds: 50\%, 75\%, and 90\%, giving three
features, giving six candidate competition features
in total.
These encode how much competing OCR evidence supports the
extraction and are computed for \hunter{} only, as the
primary extraction call.

\textit{Label and value OCR confidence} 
For each call (both \hunter{} and \mapper{}), we fuzzily match the
LLM-generated label text, extracted value, and reported nearby tokens
against the document OCR token list.
The matched OCR tokens are converted to bounding boxes in document
coordinates, and we take the minimal axis-aligned bounding box that
encloses all of them; this union box defines the attended region for
that field and call.
Within this region, we compute \textit{label and value OCR confidence}
by matching the LLM-generated label text and extracted value against
the OCR tokens using fuzzy matching (Levenshtein ratio or FuzzyWuzzy,
depending on field type), and taking the highest match confidence as
the label confidence and value confidence, respectively.
This is computed for both \hunter{} and \mapper{}, giving four
region-level confidence scores.

\paragraph{Group 3. Cross-Call Value Agreement}
\textit{Exact match}: binary indicator of
whether the \hunter{} and \mapper{} extracted values are
identical after normalisation.

\textit{Similarity match}: Levenshtein ratio
for numeric, ID, and limited-text fields; FuzzyWuzzy partial
ratio for free-text fields.
Field type governs the matching function because numeric
fields require near-exact agreement while free-text fields
tolerate surface variation.

\begin{figure}[t]
\centering
\includegraphics[width=\textwidth, trim=0 0 170pt 0, clip]{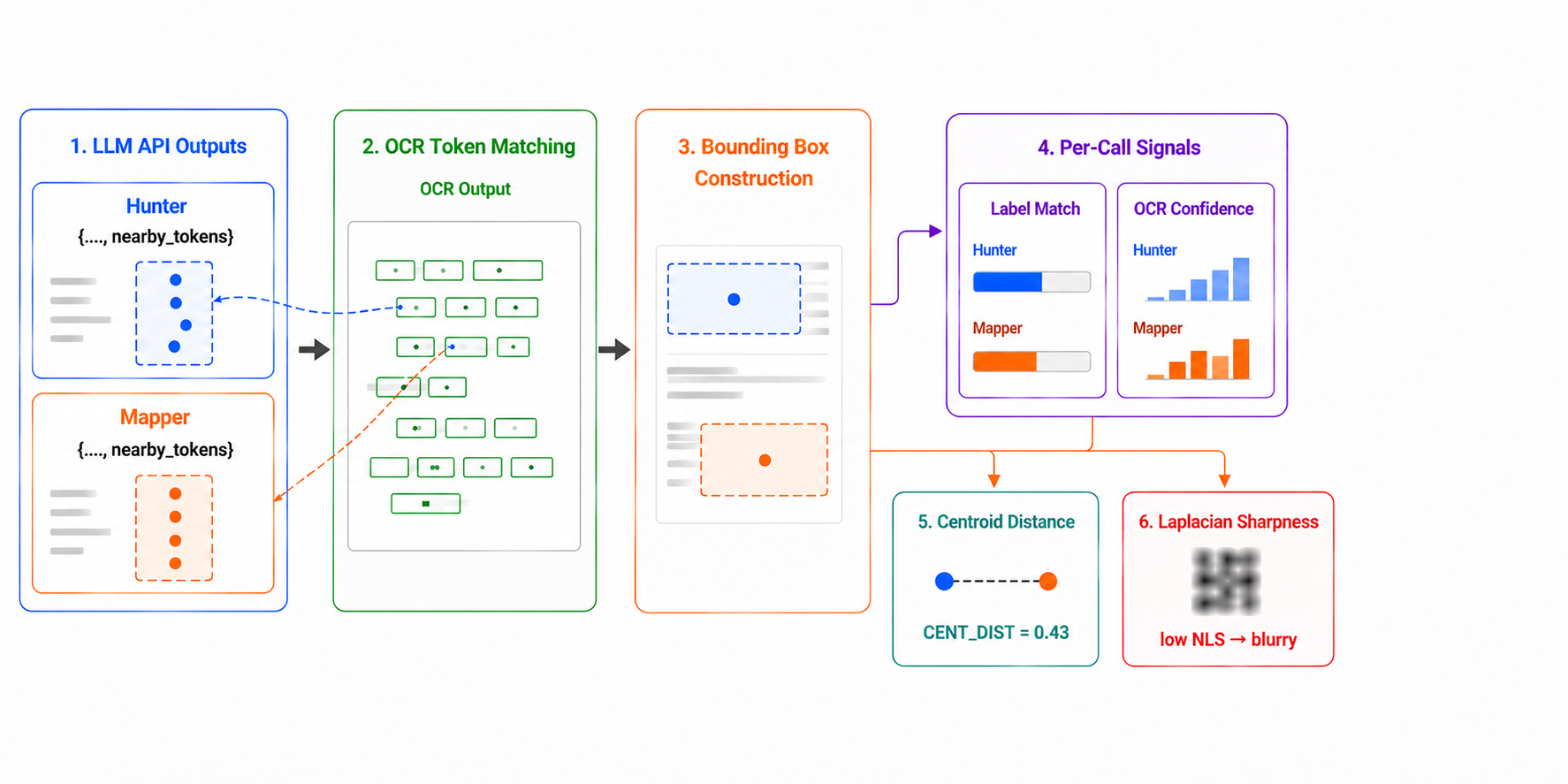}
\caption{Group 4 spatial feature extraction pipeline.
Nearby tokens from each call are matched against OCR tokens
to construct per-call bounding boxes, from which label
neighbourhood match scores, bounding box OCR confidence,
and Laplacian sharpness are derived.
Centroid distance between the two boxes measures spatial
divergence between \hunter{} and \mapper{} groundings.}
\label{fig:flowdiagram.png}
\end{figure}

\paragraph{Group 4. Spatial Neighbourhood and Centroid}
Each LLM call returns the neighbouring tokens it attended
to when generating the extraction.
We use these self-reported neighbours to construct a
spatially grounded representation of where each call grounded its attention in the document.

\textit{Bounding box construction}: the matched OCR tokens
from each call's neighbourhood are used to construct a
minimal bounding box enclosing the spatial region each call
attended to in document coordinate space.

\textit{Label neighbourhood match}: the LLM-reported nearby tokens for each
call are matched against the OCR token list within the bounding box constructed. 
The fraction of nearby tokens that can be matched to OCR
tokens in that region gives a 0--1 score for \hunter{} and
\mapper{} independently, reflecting how well each call's
attended region is grounded in the document's actual text.

\textit{Bounding box OCR confidence}:
the cumulative OCR confidence of all tokens falling within
each call's bounding box region.
This gives a spatially localised readability score for the
specific document area each call attended to, distinct
from the value-level OCR confidence in Group~2, which
measures match quality against the extracted string itself.
A low score indicates the bounding box region is poorly
recognised by the OCR engine, corroborating extraction
unreliability at the spatial level.

\textit{Centroid distance}: the
Euclidean distance between the centroid of \hunter{}'s
bounding box and the centroid of \mapper{}'s bounding box,
normalised by document dimensions.
A large centroid distance indicates the two calls grounded
their extraction in physically different regions of the
document, a strong reliability signal even when extracted
values agree, since spatial divergence suggests the
agreement may be coincidental rather than convergent.

\textit{Normalised Laplacian Score}: for the image region enclosed by each
call's bounding box, we compute the Laplacian variance of
the pixel intensities, normalised across the document.
We use the Laplacian specifically because it is a local,
second-order measure of image sharpness; it responds to
edges and fine print within the bounding box without being
influenced by the quality of the rest of the document.
Unlike global image quality metrics, this gives a
field-specific readability signal: a low Laplacian score on
the \hunter{} bounding box means the region the model was
reading was blurry or degraded, directly explaining why the
extraction may be unreliable independently of what the
logprobs report.

\paragraph{Group 5. Field Type Encoding}
Five one-hot features encoding inferred field type:
\texttt{numeric}, \texttt{date}, \texttt{limited\_text},
\texttt{free\_text}, \texttt{alphanumeric}.
Field type is inferred from the field name and extracted
value pattern.
These features enable the classifier to learn
type-conditional calibration without requiring separate
models per field type.

Full feature names and descriptions are given in
Appendix~\ref{sec:appendix_features}.

\subsection{Classifier and Post-Processing}

The 40 features are fused by a
CatBoost~\cite{prokhorenkova2018catboost} binary classifier,
selected for its native handling of missing values, calibrated
probability outputs, and strong performance in low-data regimes.
Hyperparameters are tuned per model via grid search over depth,
learning rate, and L2 regularisation, optimising validation
ROC~AUC with StratifiedGroupKFold to prevent document-level
leakage.
Best M6 configuration: depth~8, lr~0.1, l2~7, 1000 iterations.

\textbf{M7} applies isotonic regression to M6 output scores
on a held-out calibration set.
\textbf{M8} applies Lasso-regularised logistic regression as
a calibration layer.
Neither modifies CatBoost model weights.

\section{Experimental Setup}
\label{sec:setup}

\subsection{Datasets}

\paragraph{Implementation}
All \hunter{} and \mapper{} calls use GPT-4o
(\texttt{temperature=0.0}, \texttt{logprobs=True},
\texttt{top\_logprobs=5}).
OCR uses AWS Textract, providing per-word confidence scores
and bounding box coordinates fed directly into the feature
pipeline.

\paragraph{DocILE}
DocILE~\cite{simsa2023docile} comprises 6,680 real business
invoices with 55 KILE field categories spanning tax identifiers
(IBAN, BIC, VAT), addresses, dates, and line-item sub-fields.
We process 1339 fields, yielding 1,072 train and
267 test field-level samples (73.4\% positive rate;
26.6\% natural LLM failure rate).

\paragraph{CORD (zero-shot evaluation only)}
CORD~\cite{park2019cord} comprises 800 training and 100 test
receipts with 30 hierarchical field categories.
We evaluate the DocILE-trained \modelname{} on 828 CORD
field-level samples with no retraining, no feature modification,
and with CORD specific prompt.
Recalibration (M7/M8) on CORD uses 5-fold out-of-fold
cross-validation on the CORD test set only; no DocILE labels used.

\subsection{Baselines}

\textbf{B1.} Logprob mean: mean log-probability of the
extracted value tokens.
\textbf{B2.} Verbalized confidence: LLM self-assessed
0--100 integer score with anchored band descriptions.
\textbf{B3.} Self-consistency: 5 \hunter{} samples at
temperatures sampled uniformly from $[0, 1]$; agreement
fraction as confidence score.
All three collapse to near-all-positive classifiers at
threshold~0.5 (B1/B2: recall~1.000; B3: recall~0.994).

\subsection{Evaluation Metrics}

ROC~AUC (threshold-free discrimination);
PR~AUC;
ECE (Expected Calibration Error, 10-bin uniform);
Pos-ECE (ECE on predicted-positive samples);
Brier score (proper scoring rule);
AURC (area under risk-coverage curve, primary routing metric;
lower is better);
Precision, Recall, F1 at threshold~0.5.


\section{Main Results}
\label{sec:main}


\begin{table}[h]
\centering
\caption{%
Full results on DocILE.
B1--B3: model-free baselines.
M1--M8: trained CatBoost classifier (M7/M8 add recalibration).
All methods at threshold~0.5.
Best per column in \textbf{bold}.}
\label{tab:main}
\setlength{\tabcolsep}{4pt}
\renewcommand{\arraystretch}{1.22}
\begin{tabular}{llccccccc}
\toprule
  & \textbf{Method}
  & \textbf{AUC}$\uparrow$
  & \textbf{PR}$\uparrow$
  & \textbf{ECE}$\downarrow$
  & \textbf{Pos-ECE}$\downarrow$
  & \textbf{Brier}$\downarrow$
  & \textbf{AURC}$\downarrow$
  & \textbf{F1}$\uparrow$ \\
\midrule
B1 & Logprob Mean
  & 0.705 & 0.864 & 0.245 & 0.245 & 0.253 & 0.145 & 0.846$^{*}$ \\
B2 & Verbalized Conf.
  & 0.692 & 0.828 & 0.163 & 0.163 & 0.213 & 0.146 & 0.846$^{*}$ \\
B3 & Self-Consistency 5$\times$
  & 0.744 & 0.852 & 0.075 & 0.072 & 0.183 & 0.138 & 0.860$^{\dagger}$ \\
\midrule
M1 & Logprobs $+$ Entropy
  & 0.880 & 0.953 & 0.180 & 0.261 & 0.162 & 0.076 & 0.876 \\
M2 & Spatial only
  & 0.814 & 0.897 & 0.258 & 0.340 & 0.220 & 0.103 & 0.865 \\
M3 & Agreement only
  & 0.828 & 0.902 & 0.207 & 0.283 & 0.186 & 0.102 & 0.875 \\
M4 & OCR only
  & 0.896 & 0.949 & 0.197 & 0.219 & 0.141 & 0.079 & 0.896 \\
M5 & All except \mapper{}
  & 0.911 & 0.967 & 0.184 & 0.223 & 0.141 & 0.065 & 0.890 \\
M6 & \modelname{} full (\textbf{ours})
  & \textbf{0.928} & \textbf{0.974}
  & 0.199 & 0.237 & 0.136 & 0.043 & 0.914 \\
M7 & M6 $+$Isotonic Calibration
  & 0.926 & 0.973
  & \textbf{0.034} & \textbf{0.028}
  & \textbf{0.094} & 0.044 & 0.916 \\
M8 & M6 $+$Lasso Calibration
  & \textbf{0.928} & \textbf{0.974}
  & 0.048 & 0.028
  & \textbf{0.094} & \textbf{0.042} & \textbf{0.919} \\
\bottomrule
\multicolumn{9}{l}{\footnotesize
  $^{*}$All-positive: recall 1.000.\quad
  $^{\dagger}$Near-all-positive: recall 0.994.\quad
  Field-type encoding in all trained models.}
\end{tabular}
\end{table}

\begin{figure}[t]
\centering
\includegraphics[width=0.84\textwidth]{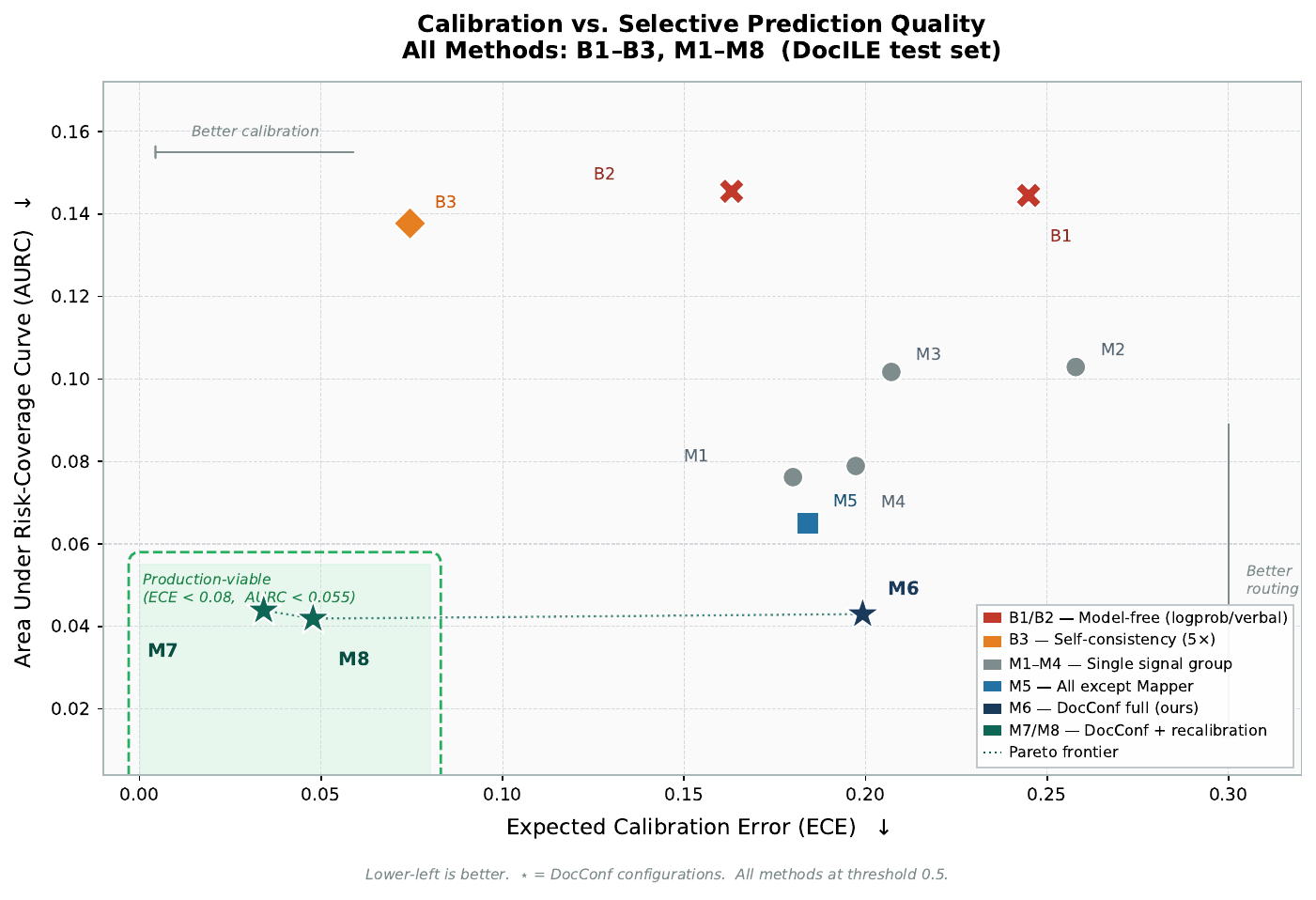}
\caption{%
\textbf{Calibration vs.\ selective prediction quality across
all methods} (DocILE test set, threshold~0.5).
Lower-left is better on both axes.
The green shaded region marks the production-viable zone
(ECE~$<$~0.08, AURC~$<$~0.055).
Baselines B1--B3 cluster in the upper-right: well-separated
from the trained models, they provide no reliable routing
signal despite B3 costing 5$\times$ more API calls.
Single-group models M1--M4 reduce AURC substantially over
baselines but remain outside the production-viable zone.
M5 crosses into viable AURC territory but is miscalibrated.
M7 and M8 are the only configurations achieving
simultaneously low ECE and low AURC, lying on the
Pareto frontier of this calibration--routing trade-off.
The dotted line connects the three \modelname{} deployment
configurations (M6, M7, M8).}
\label{fig:ece_aurc}
\end{figure}

\paragraph{(1) Self-consistency adds AUC, not routing quality}
B3 improves AUC by 3.9~pp over B1 but AURC by only 0.007.
The cheapest trained model (M1) achieves AURC~0.076,
45\% below B3, confirming that a learned continuous
function over 40 features outperforms repeated sampling of
a coarse discrete signal.

\paragraph{(2) Mapper agreement is the single largest gain}
M5$\to$M6: $+$1.7~pp AUC, $-$34\% AURC (0.065$\to$0.043),
$+$2.4~pp F1.
The cross-call agreement signal adds both discriminative and
routing quality, the largest single step in the ablation.

\paragraph{(3) OCR outperforms logprobs as a single signal group}
M4 (OCR only, 0.896~AUC) outperforms M1 (logprobs+entropy,
0.880~AUC) by 1.6~pp.
Extraction errors are document-caused, not model-caused:
OCR confidence directly measures the failure source;
logprobs measure a consequence orthogonal to it.

\paragraph{(4) Spatial features require OCR context}
M2 (Spatial only, 0.814~AUC) is the weakest single-group
model, 8.2~pp below M4.
Centroid divergence and neighbourhood overlap are informative
only when paired with the quality signal that gives them
meaning.

\paragraph{(5)  Post-hoc recalibration achieves substantially improved ECE}
M7 reduces ECE from 0.199 to 0.034 ($-$83\%) and Pos-ECE
to 0.028 at only $-$0.2~pp AUC.
M8 achieves the best AURC (0.042) with ECE~0.048.
As visible in \Cref{fig:ece_aurc}, M7 and M8 are the only
configurations in the production-viable region.

\subsection{Zero-Shot Transfer to CORD}
\label{sec:generalization}

We apply the DocILE-trained \modelname{} zero-shot to 828
CORD receipt samples, a different document type with 30 fields
vs.\ 55 and an 88.6\% positive rate, without any retraining.
Recalibration uses 5-fold OOF cross-validation on CORD only;
no DocILE supervision is transferred.

\begin{table}[h]
\centering
\caption{%
Zero-shot transfer to CORD (828 samples).
Trained on DocILE only.
Recalibration: 5-fold OOF on CORD test set.
Best per column in \textbf{bold}.}
\label{tab:generalization}
\setlength{\tabcolsep}{5pt}
\renewcommand{\arraystretch}{1.18}
\begin{tabular}{lcccccc}
\toprule
\textbf{Config}
  & \textbf{AUC}$\uparrow$
  & \textbf{ECE}$\downarrow$
  & \textbf{Brier}$\downarrow$
  & \textbf{AURC}$\downarrow$
  & \textbf{Prec.}$\uparrow$
  & \textbf{F1}$\uparrow$ \\
\midrule
B1 Base
  & \textbf{0.858} & 0.237 & 0.137 & 0.034
  & \textbf{0.954} & 0.911 \\
B2 $+$Isotonic
  & 0.839 & 0.029 & 0.081 & 0.032
  & 0.909 & 0.942 \\
B3 $+$Lasso
  & 0.854 & \textbf{0.025} & \textbf{0.078}
  & \textbf{0.029} & 0.899 & \textbf{0.940} \\
\bottomrule
\multicolumn{7}{l}{\footnotesize
  In-domain B1 ECE: 0.199.\quad
  Recalibration fitted on CORD only.}
\end{tabular}
\end{table}

\paragraph{Reliability signals are document-agnostic}
Lasso recalibration (B3) achieves ECE~0.025 on CORD,
89\% below B1's ECE~0.237, and Brier~0.078 ($-$43\%),
without any DocILE calibration labels.
The multi-signal score distribution is structurally compatible
across document types because OCR confidence, spatial centroid
divergence, and logprob entropy bear a document-agnostic
monotone relationship to extraction correctness.

\paragraph{Lasso is the recommended zero-shot configuration}
B3 achieves the best ECE (0.025), Brier (0.078), and AURC
(0.029) at only $-$0.4~pp AUC vs.\ the uncalibrated base.

\paragraph{Discrimination is preserved}
B1 achieves 0.858~AUC and 95.4\% precision zero-shot.
For deployments prioritising discrimination over calibration,
the uncalibrated base model is the optimal configuration.

\section{Analysis}
\label{sec:analysis}

\begin{figure}[t]
\centering
\begin{subfigure}[t]{0.47\linewidth}
  \centering
  \includegraphics[width=\linewidth]{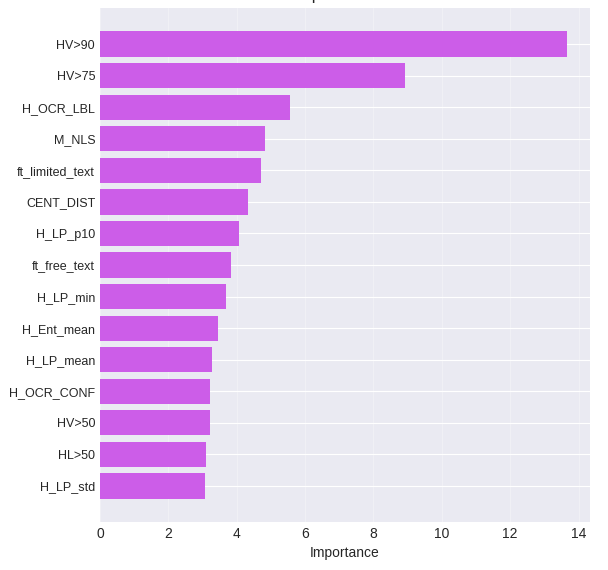}
  \caption{Top-15 CatBoost feature importances (M6).
    HV>90 and HV>75 dominate by a wide margin,
    confirming OCR value-region confidence as the primary
    correctness signal.
    M\_NLS (Mapper lexical score, rank~4) and CENT\_DIST
    (centroid divergence, rank~6) validate the cross-call
    and spatial contributions.}
  \label{fig:importance}
\end{subfigure}
\hfill
\begin{subfigure}[t]{0.47\linewidth}
  \centering
  \includegraphics[width=\linewidth,
    trim=0 0 0 0, clip]{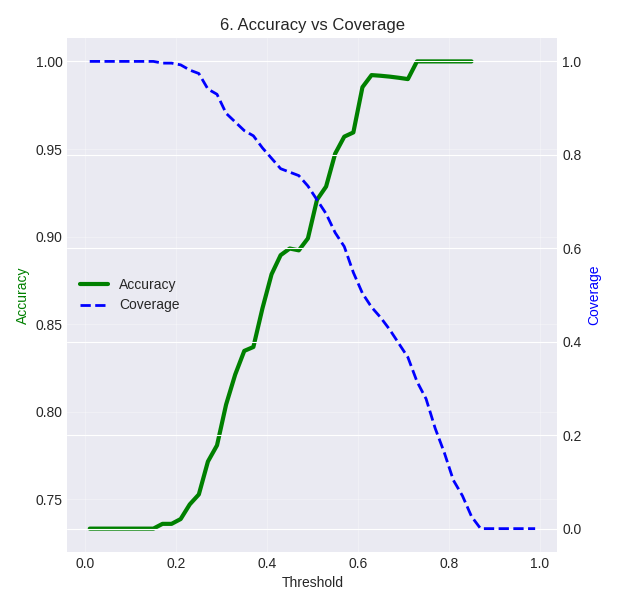}
  \caption{Cumulative accuracy vs.\ coverage (M6, DocILE
    test set).
    At 80\% coverage, routing the bottom 20\% of predictions
    to human review yields 99.1\% automated accuracy,
    a 25.8-point improvement over the 73.3\% base rate.}
  \label{fig:coverage}
\end{subfigure}
\caption{%
\textbf{(a)} Feature importances confirm that document
quality signals (HV>90, HV>75, H\_OCR\_LBL) dominate,
followed by the Mapper cross-call signal (M\_NLS),
spatial centroid divergence (CENT\_DIST), and logprob
features.
No single group is sufficient; the full fusion is required.
\textbf{(b)} The accuracy--coverage curve operationalises
the AURC metric: deferring even a modest fraction of
low-confidence predictions yields near-perfect automated
accuracy on the retained set.}
\label{fig:analysis}
\end{figure}

\subsection{Feature Importance}

\Cref{fig:analysis}a shows the top-15 CatBoost importances
from M6.
\textbf{HV>90} and \textbf{HV>75} dominate, with combined
importance nearly double the next-ranked feature.
These threshold indicators for OCR confidence on the
\hunter{}'s extracted value region explain why M4 (OCR alone,
0.896~AUC) outperforms M1 (logprobs alone, 0.880~AUC):
value-region readability is the primary correctness signal,
not model-internal uncertainty.

\textbf{M\_NLS} (rank~4) is the highest-ranked \mapper{}
feature, the normalised Laplacian score of the spatial
region \mapper{} attended to, reflecting local image quality
at the \mapper{} extraction site.
Its high rank confirms that document-side readability in
the \mapper{}'s attended region is independently predictive
of extraction correctness, separate from \hunter{}'s
OCR confidence features.

\textbf{CENT\_DIST} (rank~6) confirms spatial centroid
divergence carries independent signal: where in the document
\hunter{} and \mapper{} located the field is informative
beyond OCR quality alone.

Logprob and entropy features appear five times in the top~15,
confirming their group contribution (M1, 0.880~AUC) is real
but secondary to OCR and cross-call signals.

\subsection{Confidence--Coverage Trade-off}

\Cref{fig:analysis}b shows that at 80\% coverage,
\modelname{} achieves 99.1\% automated accuracy, a
25.8~pp improvement over the 73.3\% base rate.
This operationalises the AURC metric: a system with
AURC~0.042 delivers near-perfect accuracy at the coverage
level that matters for production automation.

\subsection{Calibration by Field Type}

Field type features (ft\_limited\_text rank~5,
ft\_free\_text rank~8) appear in the top~10, confirming
calibration quality varies across field types.
Numeric fields are well-calibrated; free-text fields show
overconfidence at high predicted probabilities.
Type-conditional routing thresholds are a direction for
future work.

\section{Discussion}
\label{sec:discussion}

\paragraph{Why document quality beats logprobs}
Extraction errors in document processing are document-caused,
not model-caused: a frontier LLM reading unreadable source
material generates high-logprob tokens about OCR noise.
OCR confidence directly measures the cause of failure;
logprobs measure a consequence orthogonal to it.
This is a structural argument applicable to any task where
LLM extraction is grounded in a physical document.

\paragraph{Why self-consistency fails for routing}
Five binary calls produce only six distinct confidence values,
making fine-grained risk stratification impossible.
Increasing to 10 calls halves the granularity gap but costs
10$\times$ per field.
\modelname{} achieves AURC~0.043 with 2 calls by learning
a continuous reliability function over 40 heterogeneous
features.

\paragraph{Why spatial signals require OCR context}
Centroid divergence and neighbourhood overlap are not
independently informative (M2, 0.814~AUC).
They require OCR quality context: spatial divergence only
signals a reliability problem when paired with evidence that
the relevant document region was difficult to read.

\paragraph{Why reliability signals transfer across domains}
Lasso recalibration achieves ECE $-$89\% on CORD without any
DocILE calibration labels.
OCR confidence, spatial centroid divergence, and logprob
entropy bear a document-agnostic monotone relationship to
extraction correctness.
A Lasso calibration layer fitted on $\sim$165 in-domain
samples appears sufficient for a new deployment domain.

\paragraph{Training data selection}
\emph{Negative sample quality dominates dataset size} for
reliability model training: a dataset with 26\% natural
failure rate provides richer negative supervision than one
with 4\%, regardless of corpus size.

\paragraph{Limitations}
\modelname{} requires explicit OCR pipeline outputs.
Evaluation covers English-language business documents.
One LLM backbone was tested; multi-backbone validation is
future work.

\section{Conclusion}
\label{sec:conclusion}

We presented \modelname{}, a reliability engine for LLM-based
document field extraction that answers the question:
\emph{should this extraction be trusted for automation?}

Five findings define the contribution.
\textit{First}, OCR value-region confidence
dominates feature importance, establishing document quality
as the primary predictor of extraction correctness.
\textit{Second}, the \hunter{}--\mapper{} dual-call design
delivers the largest single ablation gain, providing reliability signal unavailable to
any single-call or resampling approach.
\textit{Third}, \modelname{} achieves 0.928~AUC and
AURC~0.042, which is 18.4~pp above self-consistency at lower API
cost, and 70\% lower AURC than logprob-mean.
\textit{Fourth}, post-hoc recalibration reduces ECE by up
to 83\%, with three deployment configurations for different
operational priorities.
\textit{Fifth}, zero-shot Lasso recalibration on CORD reduces
ECE by 89\% and Brier by 43\% without any DocILE calibration
data, confirming genuinely document-agnostic reliability
signals.

At 80\% coverage, \modelname{} achieves 99.1\% automated
accuracy vs.\ 73.3\% base rate, enabling practical
human-in-the-loop pipelines in high-stakes document
processing.

\section{Feature Reference Table}
\label{sec:appendix_features}

Table~\ref{tab:features} provides a complete reference for
all features used in \modelname{}, grouped by signal
category.
Feature names correspond to column headers in the pipeline
output and to the feature importance plot in
Figure~\ref{fig:analysis}a.

\begingroup
\renewcommand{\arraystretch}{1.35}
\setlength{\tabcolsep}{5pt}
\begin{longtable}{p{3.2cm} p{1.4cm} p{9.2cm}}
\caption{Complete feature reference for \modelname{}.
H = \hunter{} call, M = \mapper{} call.}
\label{tab:features} \\

\toprule
\textbf{Feature Name} & \textbf{Group} & \textbf{Description} \\
\midrule
\endfirsthead

\multicolumn{3}{l}{\small\textit{Table~\ref{tab:features} continued}} \\[4pt]
\toprule
\textbf{Feature Name} & \textbf{Group} & \textbf{Description} \\
\midrule
\endhead

\midrule
\multicolumn{3}{r}{\small\textit{Continued on next page}} \\
\endfoot

\bottomrule
\endlastfoot

\multicolumn{3}{l}{\textit{Group 1 --- LLM Internal Uncertainty}} \\[2pt]
\texttt{H\_LP\_mean}   & G1 & Mean log-probability of \hunter{} output tokens \\
\texttt{H\_LP\_min}    & G1 & Minimum log-probability of \hunter{} output tokens \\
\texttt{H\_LP\_p10}    & G1 & 10th-percentile log-probability (\hunter{}) \\
\texttt{H\_LP\_std}    & G1 & Standard deviation of \hunter{} log-probabilities \\
\texttt{H\_Ent\_mean}  & G1 & Mean Shannon entropy of \hunter{} output tokens \\
\texttt{H\_Ent\_max}   & G1 & Maximum Shannon entropy (\hunter{}) \\
\texttt{H\_Ent\_p90}   & G1 & 90th-percentile Shannon entropy (\hunter{}) \\
\texttt{MV\_LP\_mean}  & G1 & Mean log-probability of \mapper{} output tokens \\
\texttt{MV\_LP\_min}   & G1 & Minimum log-probability of \mapper{} output tokens \\
\texttt{MV\_LP\_p10}   & G1 & 10th-percentile log-probability (\mapper{}) \\
\texttt{MV\_LP\_std}   & G1 & Standard deviation of \mapper{} log-probabilities \\
\texttt{MV\_Ent\_mean} & G1 & Mean Shannon entropy of \mapper{} output tokens \\
\texttt{MV\_Ent\_max}  & G1 & Maximum Shannon entropy (\mapper{}) \\
\texttt{MV\_Ent\_p90}  & G1 & 90th-percentile Shannon entropy (\mapper{}) \\[4pt]

\multicolumn{3}{l}{\textit{Group 2 --- OCR and Value Confidence}} \\[2pt]
\texttt{H\_OCR\_LBL}  & G2 & OCR confidence on the \hunter{} field label region \\
\texttt{H\_OCR\_VAL}  & G2 & OCR confidence on the \hunter{} extracted value region \\
\texttt{M\_OCR\_LBL}  & G2 & OCR confidence on the \mapper{} field label region \\
\texttt{M\_OCR\_VAL}  & G2 & OCR confidence on the \mapper{} extracted value region \\
\texttt{HL>50}        & G2 & Fraction of \hunter{} label-region OCR candidates with similarity $\geq$50\% to extracted label \\
\texttt{HL>75}        & G2 & Fraction of \hunter{} label-region OCR candidates with similarity $\geq$75\% to extracted label \\
\texttt{HL>90}        & G2 & Fraction of \hunter{} label-region OCR candidates with similarity $\geq$90\% to extracted label \\
\texttt{HV>50}        & G2 & Fraction of \hunter{} value-region OCR candidates with similarity $\geq$50\% to extracted value \\
\texttt{HV>75}        & G2 & Fraction of \hunter{} value-region OCR candidates with similarity $\geq$75\% to extracted value \\
\texttt{HV>90}        & G2 & Fraction of \hunter{} value-region OCR candidates with similarity $\geq$90\% to extracted value \\
\texttt{H\_VAL\_EX}   & G2 & Binary: \hunter{} value found anywhere in raw OCR text of the document \\
\texttt{M\_VAL\_EX}   & G2 & Binary: \mapper{} value found anywhere in raw OCR text of the document \\[4pt]

\multicolumn{3}{l}{\textit{Group 3 --- Cross-Call Value Agreement}} \\[2pt]
\texttt{EX}   & G3 & Binary: \hunter{} and \mapper{} extracted values are identical after normalisation \\
\texttt{SIM}  & G3 & Levenshtein ratio (numeric/ID/limited-text fields) or FuzzyWuzzy partial ratio (free-text fields) between \hunter{} and \mapper{} values \\[4pt]

\multicolumn{3}{l}{\textit{Group 4 --- Spatial Neighbourhood and Centroid}} \\[2pt]
\texttt{H\_LBLS}    & G4 & Fraction of \hunter{} nearby tokens matched to OCR tokens in the defined region of interest (0--1) \\
\texttt{M\_LBLS}    & G4 & Fraction of \mapper{} nearby tokens matched to OCR tokens in the defined region of interest (0--1) \\
\texttt{H\_OCR\_CONF} & G4 & Cumulative OCR confidence of all tokens within the \hunter{} bounding box region \\
\texttt{M\_OCR\_CONF} & G4 & Cumulative OCR confidence of all tokens within the \mapper{} bounding box region \\
\texttt{CENT\_DIST} & G4 & Euclidean distance between \hunter{} and \mapper{} bounding box centroids, normalised by document dimensions \\
\texttt{H\_NLS}     & G4 & Normalised Laplacian score of the image region enclosed by \hunter{}'s bounding box; measures local sharpness at the extraction site \\
\texttt{M\_NLS}     & G4 & Normalised Laplacian score of the image region enclosed by \mapper{}'s bounding box; measures local sharpness at the extraction site \\[4pt]

\multicolumn{3}{l}{\textit{Group 5 --- Field Type Encoding}} \\[2pt]
\texttt{ft\_numeric}       & G5 & One-hot: field type inferred as numeric \\
\texttt{ft\_date}          & G5 & One-hot: field type inferred as date \\
\texttt{ft\_limited\_text} & G5 & One-hot: field type inferred as limited text (short alphanumeric strings) \\
\texttt{ft\_free\_text}    & G5 & One-hot: field type inferred as free text (longer natural language values) \\
\texttt{ft\_alphanumeric}  & G5 & One-hot: field type inferred as alphanumeric (codes, identifiers) \\

\end{longtable}
\endgroup
\end{document}